# An Extended Neo-Fuzzy Neuron and its Adaptive Learning Algorithm

**Yevgeniy V. Bodyanskiy**
Kharkiv National University of Radio Electronics, Kharkiv, Ukraine,
Email: bodya@kture.kharkov.ua

**Oleksii K. Tyshchenko and Daria S. Kopaliani**
Kharkiv National University of Radio Electronics, Kharkiv, Ukraine,
Email: { lehatish, daria.kopaliani }@gmail.com

*Abstract* — A modification of the neo-fuzzy neuron is proposed (an extended neo-fuzzy neuron (ENFN)) that is characterized by improved approximating properties. An adaptive learning algorithm is proposed that has both tracking and smoothing properties and solves prediction, filtering and smoothing tasks of non-stationary "noisy" stochastic and chaotic signals. An ENFN distinctive feature is its computational simplicity compared to other artificial neural networks and neuro-fuzzy systems.

## I. INTRODUCTION

Artificial neural networks (ANNs) are currently widely used for solving different Data Mining tasks due to their approximating capabilities and their ability to learn from experimental data [1-3]. However, when the data come sequentially in real time, many neural networks lose their effectiveness because of the multiepoch learning (which is used in many ANNs and designated only for a batch mode). Of course, radial-basis-function networks (RBFN) could be used in such situations. These ANNs are characterized by a high speed of learning processes, but, first of all, these networks suffer from the so-called «curse of dimensionality» and, secondly, even a trained neural network is a «black box», and its results can not be interpreted. Hybrid systems of computational intelligence [5-7], and above all neuro-fuzzy systems (NFSs), combining the advantages of ANNs and fuzzy inference systems (FISs), do not suffer from the "curse of dimensionality" and provide linguistic interpretability and transparency of the results. However, since most of the well-known NFSs are trained with the help of the error backpropagation concept, they are ill-equipped to work in an online mode.

Due to the above mentioned problems, we would like to develop hybrid systems of computational intelligence that deal with processing the incoming data in an online mode and have advantages of both ANNs and NFSs.

The remainder of this paper is organized as follows: Section 2 gives a neo-fuzzy neuron architecture. Section 3 describes an extended neo-fuzzy neuron architecture. Section 4 presents experiments and evaluation. Conclusions and future work are given in the final section.

## II. THE NEO-FUZZY NEURON ARCHITECTURE

To overcome the above mentioned problems, a neuro-fuzzy system called by the authors a "neo-fuzzy neuron (NFN)" was introduced in [8 – 10]. The architecture of the neo-fuzzy neuron is in fig.1.

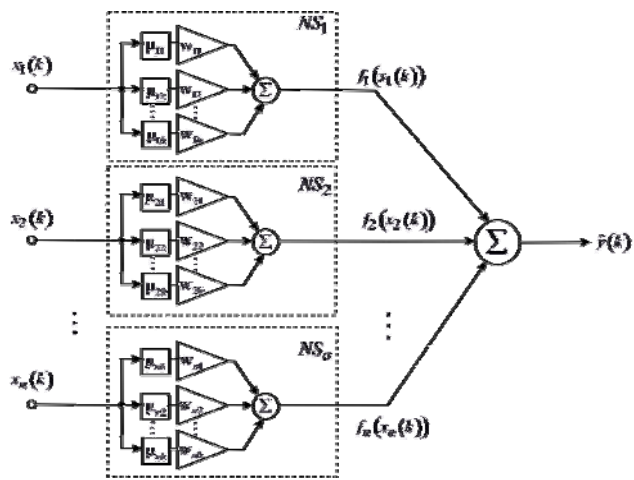

Fig.1. A neo-fuzzy neuron

The neo-fuzzy neuron is a nonlinear learning system with multiple inputs and one output that implements the mapping

$$\hat{y} = \sum_{i=1}^{n} f_i(x_i) \qquad (1)$$

where $x_i$ is the $i$-th component of the $n$-dimensional input signal vector $x = (x_1,...,x_i,...,x_n)^T \in R^n$, $\hat{y}$ is a scalar NFN output. The NFN structural blocks are nonlinear synapses $NS_i$ that carry out a nonlinear transformation of the $i$-th component of $x_i$ in the form

$$f_i(x_i) = \sum_{l=1}^{h} w_{li} \mu_{li}(x_i) \qquad (2)$$

where $w_{li}$ is the $l$-th synaptic weight of the $i$-th nonlinear synapse, $l = 1,2,...,h$, $i = 1,2,...,n$; $\mu_{li}(x_i)$ is the $l$-th membership function in the $i$-th nonlinear synapse performing fuzzification of a crisp component $x_i$. Thus, transformation implemented by the NFN can be written as

$$\hat{y} = \sum_{i=1}^{n} \sum_{l=1}^{h} w_{li} \mu_{li}(x_i). \qquad (3)$$

Fuzzy inference implemented by the same NFN has the form
IF $x_i$ IS $X_{li}$ THEN THE OUTPUT IS

$$w_{li}, \; l = 1,2,...,h, \qquad (4)$$

which means that a nonlinear synapse actually implements the zero-order Takagi-Sugeno fuzzy inference [11, 12].

The NFN authors [8–10] used traditional triangular constructions as membership functions that meet the unity partition conditions

$$\mu_{li}(x_i) = \begin{cases} \dfrac{x_i - c_{l-1,i}}{c_{li} - c_{l-1,i}} & \text{if } x_i \in [c_{l-1,i}, c_{li}], \\ \dfrac{c_{l+1,i} - x_i}{c_{l+1,i} - c_{li}} & \text{if } x_i \in [c_{li}, c_{l+1,i}], \\ 0 & \text{otherwise} \end{cases} \qquad (5)$$

where $c_{li}$ stands for arbitrarily selected (usually uniformly distributed) centers of the membership functions in the interval [0, 1], thus, naturally $0 \le x_i \le 1$.

Such a choice of the membership functions leads to the fact that the $i$-th component of the input signal $x_i$ activates only two adjacent functions, thus their sum is equal to 1 which means that

$$\mu_{li}(x_i) + \mu_{l+1,i}(x_i) = 1 \qquad (6)$$

and

$$f_i(x_i) = w_{li} \mu_{li}(x_i) + w_{l+1,i} \mu_{l+1,i}(x_i). \qquad (7)$$

Just exactly this circumstance allows to synthesize simple and effective adaptive controllers for nonlinear control objects [13, 14].

One can use other membership functions except triangular constructions, first of all, B-splines [15] that proved their effectiveness in the neo-fuzzy neurons [16]. A general case of the membership functions based on the $q$-th degree B-spline can be presented in the form

$$\mu_{li}^B(x_i, q) = \begin{cases} \begin{cases} 1 \text{ if } x_i \in [c_{li}, c_{l+1,i}] \\ 0 \text{ otherwise} \end{cases} \text{ for } q = 1, \\ \dfrac{x_i - c_{li}}{c_{l+q-1,i} - c_{li}} \mu_{li}^B(x_i, q-1) + \\ + \dfrac{c_{l+q,i} - x_i}{c_{l+q,i} - c_{l+1,i}} \mu_{l+1,i}^B(x_i, q-1) \text{ for } q > 1, \\ l = 1, 2,..., h\text{-}q. \end{cases} \qquad (8)$$

When $q = 2$, we obtain the traditional triangular functions. It should be mentioned that B-splines also provide the unity partition in the form

$$\sum_{l=1}^{h} \mu_{li}^B(x_i, q) = 1,$$

they are non-negative which means that

$$\mu_{li}^{B}(x_{i},q) \geq 0$$

and have a local support

$$\mu_{li}^{B}(x_{i},q) = 0 \text{ for } x_{i} \notin \left[c_{li},c_{l+q,i}\right].$$

When the vector signal $x(k)=(x_{1}(k),...,x_{i}(k),...,x_{n}(k))^{T}$ ($k=1,2,...$ – the current discrete time) is fed to the NFN input, a scalar value is calculated at the NFN output

$$\hat{y}(k) = \sum_{i=1}^{n}\sum_{l=1}^{h} w_{li}(k-1)\mu_{li}(x_{i}(k)) \qquad (9)$$

where $w_{li}(k-1)$ is the current value of adjusted synaptic weights that were obtained from a learning procedure of the previous $(k-1)$ observations.

Introducing a $(nh \times 1)$ – membership functions vector $\mu(x(k)) = (\mu_{11}(x_{1}(k)),...,\mu_{h1}(x_{1}(k)),\mu_{12}(x_{2}(k)),...,$
$\mu_{li}(x_{i}(k)),...,\mu_{hn}(x_{n}(k)))^{T}$ and a corresponding synaptic weights vector
$w(k-1) = (w_{11}(k-1),...,w_{h1}(k-1),w_{12}(k-1),...,w_{li}(k-1),...,w_{hn}(k-1))^{T}$, we can write the transformation (9) implemented by the NFN in a compact form

$$\hat{y}(k) = w^{T}(k-1)\mu(x(k)). \qquad (10)$$

To adjust the neo-fuzzy neuron parameters, the authors used a gradient procedure that minimizes the learning criterion

$$E(k) = \frac{1}{2}(y(k)-\hat{y}(k))^{2} = \frac{1}{2}e^{2}(k) =$$
$$= \frac{1}{2}\left(y(k) - \sum_{i=1}^{n}\sum_{l=1}^{h} w_{li}\mu_{li}(x_{i}(k))\right)^{2} \qquad (11)$$

and has the form

$$w_{li}(k) = w_{li}(k-1) + \eta e(k)\mu_{li}(x_{i}(k)) =$$
$$= w_{li}(k-1) + \eta(y(k)-\hat{y}(k))\mu_{li}(x_{i}(k)) =$$
$$= w_{li}(k-1) + \qquad (12)$$
$$+ \eta\left(y(k) - \sum_{i=1}^{n}\sum_{l=1}^{h} w_{li}(k-1)\mu_{li}(x_{i}(k))\right)\mu_{li}(x_{i}(k))$$

where $y(k)$ is an external reference signal, $e(k)$ is a learning error, $\eta$ is a learning rate parameter.

A special algorithm was proposed in [17] to accelerate the NFN learning procedure which has both tracking (for non-stationary signal processing) and filtering properties (for "noisy" data processing)

$$\begin{cases} w(k) = w(k-1) + r^{-1}(k)e(k)\mu(x(k)), \\ r(k) = \alpha r(k-1) + \|\mu(x(k))\|^{2}, 0 \leq \alpha \leq 1. \end{cases} \qquad (13)$$

When $\alpha = 0$, the algorithm (13) is identical to the one-step Kaczmarz-Widrow-Hoff learning algorithm [18] and when $\alpha = 1$ – to the Goodwin-Ramage-Caines stochastic approximation algorithm [19].

It should be mentioned that one can use many other learning and identification algorithms including the traditional least-squares method with all modifications to train the NFN synaptic weights.

III. AN EXTENDED NEO-FUZZY NEURON

As mentioned above, the NFN nonlinear synapse $NS_{i}$ implements the zero-order Takagi-Sugeno inference thus being the elementary Wang-Mendel neuro-fuzzy system [20–22]. It is possible to improve approximating properties of such a system by using a special structural unit which we called an "extended nonlinear synapse" ($ENS_{i}$, fig.2) and to synthesize an "extended neo-fuzzy neuron" (ENFN) that contains $ENS_{i}$ elements instead of usual nonlinear synapses $NS_{i}$ (fig.3).

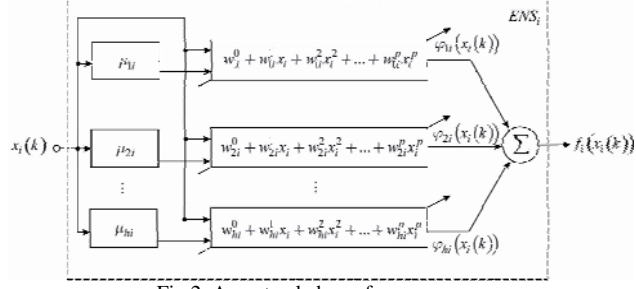
Fig.2. An extended neo-fuzzy synapse

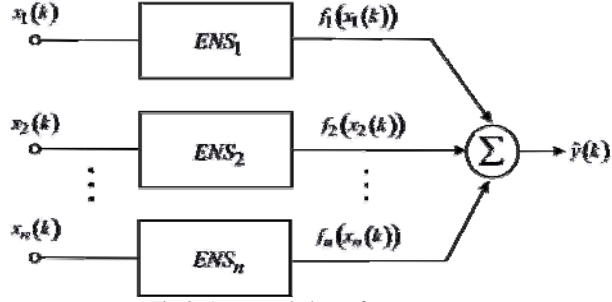
Fig.3. An extended neo-fuzzy neuron

Introducing additional variables

$$\varphi_{li}(x_i) = \mu_{li}(x_i)\left(w_{li}^0 + w_{li}^1 x_i + w_{li}^2 x_i^2 + \ldots + w_{li}^p x_i^p\right), \quad (14)$$

$$\begin{aligned}f_i(x_i) &= \sum_{l=1}^{h} \mu_{li}(x_i)\left(w_{li}^0 + w_{li}^1 x_i + w_{li}^2 x_i^2 + \ldots + w_{li}^p x_i^p\right) = \\ &= w_{1i}^0 \mu_{1i}(x_i) + w_{1i}^1 x_i \mu_{1i}(x_i) + \ldots + w_{1i}^p x_i^p \mu_{1i}(x_i) + \\ &\quad + w_{2i}^0 \mu_{2i}(x_i) + \ldots + w_{2i}^p x_i^p \mu_{2i}(x_i) + \ldots + w_{hi}^p x_i^p \mu_{hi}(x_i),\end{aligned} \quad (15)$$

$$w_i = \left(w_{1i}^0, w_{1i}^1, \ldots, w_{1i}^p, w_{2i}^0, \ldots, w_{2i}^p, \ldots, w_{hi}^p\right)^T, \quad (16)$$

$$\tilde{\mu}_i(x_i) = \bigl(\mu_{1i}(x_i), x_i \mu_{1i}(x_i), \ldots, x_i^p \mu_{1i}(x_i), \\ \mu_{2i}(x_i), \ldots, x_i^p \mu_{2i}(x_i), \ldots, x_i^p \mu_{hi}(x_i)\bigr)^T, \quad (17)$$

it can be written

$$f_i(x_i) = w_i^T \tilde{\mu}_i(x_i), \quad (18)$$

$$\hat{y} = \sum_{i=1}^{n} f_i(x_i) = \sum_{i=1}^{n} w_i^T \tilde{\mu}(x_i) = \tilde{w}^T \tilde{\mu}(x) \quad (19)$$

where $\tilde{w}^T = \left(w_1^T, \ldots, w_i^T, \ldots, w_n^T\right)^T$, $\tilde{\mu}(x) = \left(\tilde{\mu}_1^T(x_1), \ldots, \tilde{\mu}_i^T(x_i), \ldots, \tilde{\mu}_n^T(x_n)\right)^T$.

It's easy to see that the ENFN contains $(p+1)hn$ adjusted synaptic weights and the fuzzy inference implemented by each $ENS_i$ has the form

$$\begin{aligned}&\text{IF } x_i \text{ IS } X_{li} \text{ THEN THE OUTPUT IS} \\ &w_{li}^0 + w_{li}^1 x_i + \ldots + w_{li}^p x_i^p, \ l = 1, 2, \ldots, h\end{aligned} \quad (20)$$

which coincides with the $p$-order Takagi-Sugeno inference.

The ENFN has a much simpler architecture than a traditional neuro-fuzzy system that simplifies its numerical implementation.

When the vector signal $x(k)$ is fed to the ENFN input, a scalar value is calculated at the ENFN output

$$\hat{y}(k) = \tilde{w}^T(k-1)\tilde{\mu}(x(k)) \quad (21)$$

wherein this expression differs from the expression (10) by the fact that it contains $(p+1)$ times more adjusted parameters than the conventional NFN. It is clear that the algorithm (13) may be used for training ENFN parameters obtaining the form in this case

$$\begin{cases} \tilde{w}(k) = \tilde{w}(k-1) + \tilde{r}^{-1}(k)e(k)\tilde{\mu}(x(k)), \\ \tilde{r}(k) = \alpha\tilde{r}(k-1) + \|\tilde{\mu}(x(k))\|^2, 0 \leq \alpha \leq 1. \end{cases} \quad (22)$$

IV. EXPERIMENT AND ANALYSIS

To demonstrate the efficiency of the proposed adaptive neuro-fuzzy system and its learning procedure (22), we have implemented a simulation test based on forecasting of a chaotic process defined by the Mackey-Glass equation [23]

$$y'(t) = \frac{0.2t(t-\tau)}{1+y^{10}(t-\tau)} - 0.1y(t). \quad (23)$$

The signal defined by (22) was quantized with a step 0.1. We took a fragment containing 12000 points. The goal was to predict a time-series value on the next step $k+1$ using its values on steps $k-3$, $k-2$, $k-1$, and $k$.

First 7000 points were used as a training set (for adjusting weight coefficients of the architecture), next 5000 points were used as a test set (7001-12000) without adjusting weight coefficients. $p$ is a fuzzy inference order, a number of membership functions $h$ is 3, a smoothing parameter $\alpha$ is 0.9 during the weight adaptation procedure in (21).

We implement one step prediction in all our experiments.

Symmetric mean absolute percentage error (SMAPE), root mean square error (RMSE) and mean square error (MSE), used for result evaluation, are shown in Tab.1-5.

Fig.4-8 present time series outputs, prediction values and prediction errors (a time series value is marked with a blue color, a prediction value is marked with a green color, a prediction error is marked with a red color).

The proposed algorithm gives the close approximation and the high prediction quality of sufficiently non-stationary processes in an online mode.

Table 1. Prediction results of the Mackey-Glass time series

|     | RMSEtest  | MSEtest   | SMAPEtest |
| --- | --------- | --------- | --------- |
| p=0 | 0.0105742 | 0.0022596 | 7.4159348 |
| p=1 | 0.0064418 | 0.0004191 | 3.5964145 |
| p=2 | 0.0007427 | 0.0003537 | 3.3670534 |
| p=3 | 0.0001568 | 0.0004181 | 3.6279506 |
| p=5 | 0.0009585 | 0.0005421 | 4.0177900 |

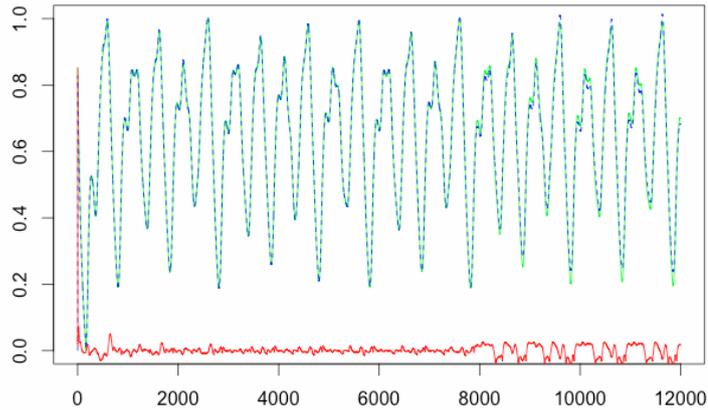

Fig.4. The Mackey-Glass time series prediction
(p = 3, h = 3, $\alpha$ = 0.9).

The first Narendra object [24] is the dynamic plant identification problem and is described by the equation

$$y(k+1) = y(k)/(1+y^2(k)) + f(k) \quad (24)$$

where

$$f(k) = \begin{cases} \sin^3(\pi k / 250) & \text{if } k > 500, \\ 0.8\sin(\pi k / 250) + 0.2\sin(\pi k / 25) \\ \text{otherwise.} \end{cases}$$

Table 2. Prediction results of the Narendra object

|     | RMSEtest  | MSEtest   | SMAPEtest |
|-----|-----------|-----------|-----------|
| p=0 | 0.0025481 | 0.0021746 | 18.347822 |
| p=1 | 0.0051009 | 0.0026075 | 11.220434 |
| p=2 | 0.0012913 | 0.0012601 | 5.5851695 |
| p=3 | 0.0009847 | 0.0011488 | 5.5871958 |
| p=5 | 0.0006739 | 0.0010488 | 5.5533630 |

A generated sequence contains 2000 values. We used $f(k) = \sin^3(\pi k / 250)$ for the first 500 points (a training set, $k = 1...500$) and $f(k) = 0.8\sin(\pi k / 250) + 0.2\sin(\pi k / 25)$ for a test set ($k = 501...2000$).

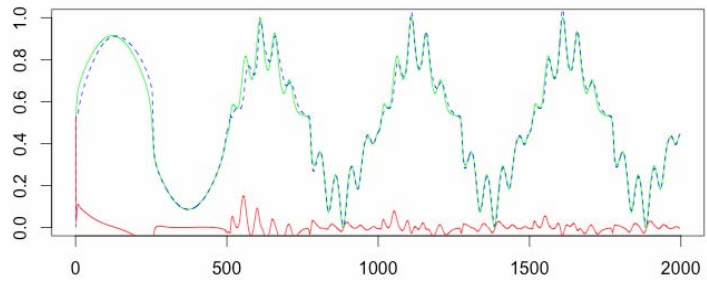

Fig.5. The Narendra time series prediction
(p = 3, h = 3, $\alpha = 0.9$).

The second Narendra plant is assumed to be in the form

$$y(k+3) = f(y(k+2), y(k+1), y(k), u(k+3), u(k+2)) \quad (25)$$

where

$$f(x_1, x_2, x_3, x_4, x_5) = (x_1 x_2 x_4 x_5 (x_3 - 1) + x_4) / (1 + x_3^2 + x_2^2).$$

A generated sequence contains 1500 values. The input to the plant is given by $u(k) = \sin(\pi k / 25)$ for $k < 250$, $u(k) = 1$ for $250 \leq k \leq 500$, $u(k) = -1$ for $501 \leq k \leq 750$ and $u(k) = 0.4\sin(\pi k / 25) + 0.1\sin(\pi k / 32) + 0.6\sin(\pi k / 10)$ for $k > 751$.

Table 3. Prediction results of the Narendra object

|     | RMSEtest  | MSEtest   | SMAPEtest  |
|-----|-----------|-----------|------------|
| p=0 | 0.0024681 | 0.0067067 | 14.5441772 |
| p=1 | 0.0049146 | 0.0054865 | 12.4697803 |
| p=2 | 0.0017915 | 0.0023304 | 9.3855439  |
| p=3 | 0.0015882 | 0.0023559 | 9.4933733  |
| p=5 | 0.0014513 | 0.0024112 | 9.5267278  |

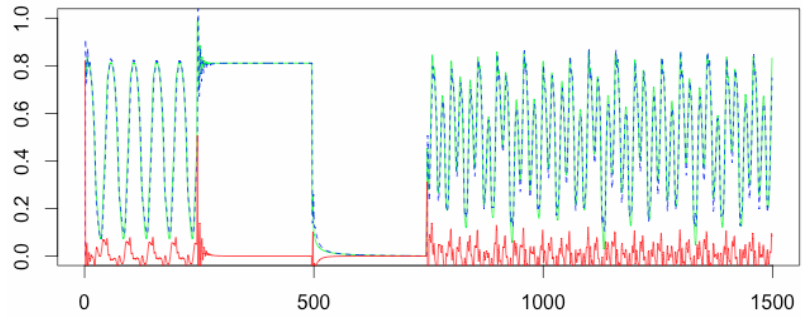

Fig.6. The Narendra time series prediction
(p = 3, h = 3, $\alpha$ = 0.9).

The third Narendra object is described by the equation

$$y(k+1) = y(k)/(1+y^2(k)+f(k)) \quad (26)$$

where

$$f(k) = (\cos(2\pi k/25)+\cos(2\pi k/2))^3 \text{ if } k < 2000,$$
$$(\sin(2\pi k/250)+\sin(2\pi k/10))^3 \text{ otherwise}$$

A generated sequence contains 4000 values.

Table 4. Prediction results of the Narendra object

|  | RMSEtest | MSEtest | SMAPEtest |
|---|---|---|---|
| p=0 | 0.0015450 | 0.0012522 | 11.3478218 |
| p=1 | 0.0028319 | 0.0016039 | 14.2204343 |
| p=2 | 0.0010964 | 0.0007985 | 6.4813514 |
| p=3 | 0.0008922 | 0.0007470 | 6.5637786 |
| p=5 | 0.0006855 | 0.0007125 | 6.6079069 |

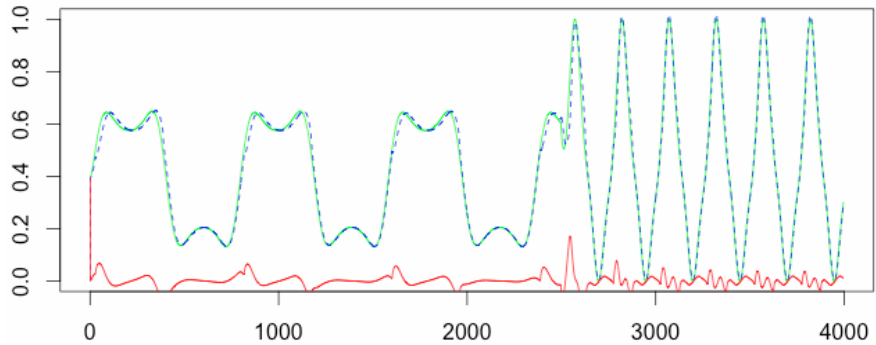

Fig.7. The Narendra time series prediction
(p = 5, h = 3, $\alpha$ = 0.9).

The forth Narendra object is described in the form

$$y(k+1) = y(k)/(1+y^2(k)) + \sin(2\pi k/25) + \\ +\sin(2\pi k/10) \quad (27)$$

A generated sequence contains 500 values.

Table 5. Prediction results of the Narendra object

|  | RMSEtest | MSEtest | SMAPEtest |
|---|---|---|---|
| p=0 | 0.0080359 | 0.0214440 | 25.1994989 |
| p=1 | 0.0114304 | 0.0194163 | 25.3300074 |

| p=2 | 0.0017572 | 0.0104677 | 20.2466940 |
|---|---|---|---|
| p=3 | 0.0010388 | 0.0106037 | 20.2904977 |
| p=5 | 8.6533285 | 0.0109404 | 20.4180131 |

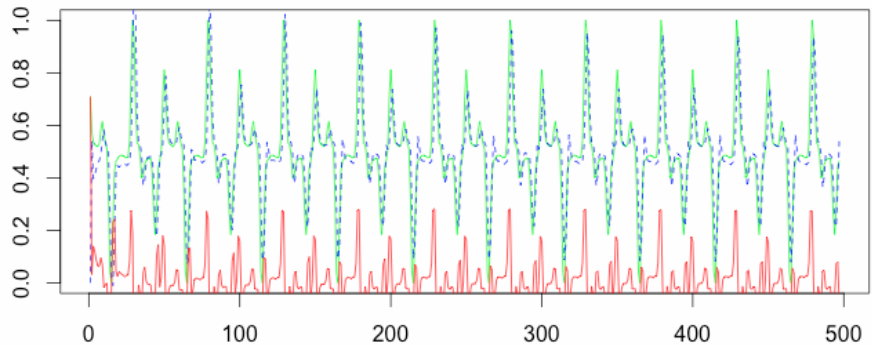

Fig.8. The Narendra time series prediction
(p = 5, h = 3, $\alpha$ = 0.9).

As it can be seen from the fig.4-8, the ENFN approximating properties are better when compared to the traditional NFN architecture which is in fact an ENFN special case (the NFN implements the zero-order Takagi-Sugeno fuzzy inference, p=0).

## VI. CONCLUSION

The architecture of an extended neo-fuzzy neuron is proposed in the paper which is a generalization of the standard neo-fuzzy neuron in a case of the "above zero"-order fuzzy inference. The learning algorithm is proposed that is characterized by both tracking and filtering properties. The extended NFN has improved approximating properties, it's characterized by a high learning rate, it also has simple numerical implementation.

## ACKNOWLEDGMENT

The authors would like to thank anonymous reviewers for their careful reading of this paper and for their helpful comments.